\documentclass{sig-alternate}

\usepackage[numbers]{natbib}

\makeatletter
\def\@copyrightspace{\relax}
\makeatother

\usepackage{fancyhdr}
\begin{document}
\sloppy  





%

\title{Framing Effects on Privacy Concerns \\about a Home Telepresence Robot}


\numberofauthors{1}
\author{
  \alignauthor
  Matthew Rueben$^1$, Frank J. Bernieri$^2$, Cindy M. Grimm$^1$, \\and William D. Smart$^1$\\
    \affaddr{$^1$Robotics Program}\\
    \affaddr{$^2$School of Psychological Science}\\
    \affaddr{Oregon State University}\\
    \affaddr{Corvallis, OR 97331}\\
    \affaddr{United States}\\
    \affaddr{\email{\{ruebenm,frank.bernieri,cindy.grimm,bill.smart\}@oregonstate.edu}}
}

\date{19 December 2016}

\CopyrightYear{2017} 
\setcopyright{acmcopyright}
\conferenceinfo{HRI '17,}{March 06-09, 2017, Vienna, Austria}
\isbn{978-1-4503-4336-7/17/03}\acmPrice{\$15.00}
\doi{http://dx.doi.org/10.1145/2909824.3020218}

\maketitle
\begin{abstract}
Privacy-sensitive robotics is an emerging area of HRI research. Judgments about privacy would seem to be context-dependent, but none of the promising work on contextual ``frames'' has focused on privacy concerns. This work studies the impact of contextual ``frames'' on local users' privacy judgments in a home telepresence setting. Our methodology consists of using an online questionnaire to collect responses to animated videos of a telepresence robot after framing people with an introductory paragraph. 

The results of four studies indicate a large effect of manipulating the robot operator's identity between a stranger and a close confidante. It also appears that this framing effect persists throughout several videos. These findings serve to caution HRI researchers that a change in frame could cause their results to fail to replicate or generalize. We also recommend that robots be designed to encourage or discourage certain frames.



%
%
%
%
%
\end{abstract}


\section{Introduction}
\label{sec:Introduction}
Unique to HRI research is the human's interpretation of a scenario; human perceptions and behaviors around robots are unpredictable given only the external, physical facts about the scenario. This work focuses on the impact of frames---i.e., ``structure[s] of expectation" \citep{tannen_whats_1993} within which actions and words will be interpreted differently. It is our intuition that the frame surrounding a given interaction could have comparable or even larger effects on judgments about that interaction than the independent variables typically studied in HRI research, e.g., robot morphology, behavior, environmental factors, and individual differences between subjects. We suspect that even a well-designed HRI can make a bad impression if it is framed such that observers interpret the robot's behaviors negatively; on the other hand, understanding framing effects might be a more efficient way than modifying robot appearance and behavior for reducing or reversing negative reactions to robots. \citet{groom_responses_2011} observe that ``[HRI] researchers have largely ignored studying framing as an independent variable.'' We seek to reverse this trend: the studies presented in this paper use framing to manipulate each subject's relationship with the robot operator via short operator biographies in (uniquely) a telepresence scenario. 

Our study of framing effects is motivated by privacy concerns in HRI. Privacy is important in all human cultures \citep{altman_privacy_1977}, although different cultures have different norms for privacy and different mechanisms for enforcing those norms. We use the word ``privacy'' to describe a bundle of constructs related to perceived control over informational, physical, psychological, and social aspects of one's life \citep{rueben_taxonomy_2016}. It seems clear that telepresence robots, like autonomous robots, cause concerns about privacy. Telepresence robots are essentially video media spaces, which have a slew of privacy problems themselves (see \citet{boyle_privacy_2009} for a review), but also mobile, which adds new privacy concerns. Telepresence robots can be driven into private spaces, or used to look around at things against the will of the local user(s). Our broad research focus is on how privacy judgments work in robot-mediated communication; we suspect that framing is among the main factors.

The goal of our research is to answer the question, ``how does framing impact privacy judgments?'' We have run four human-subjects experiments in rapid succession to gather the first measurements of these effects. Our approach is to use some text to frame a scenario presented via an animated video, to which subjects respond via a questionnaire. We recruit subjects using Amazon Mechanical Turk for quick development and turnaround, beginning with simpler scenarios and variables and then progressing to the privacy concerns that motivate our research.

Each study builds on the previous ones. \emph{Study 1} tests whether we can measure a framing effect with our approach. \emph{Study 2} expands upon the first one to include the effect of re-framing people partway through the study. \emph{Study 3} tests whether our findings generalize to a new video and whether different demographic subgroups are differentially affected by framing. \emph{Study 4} uses a suite of high-fidelity animations specifically designed to evoke privacy concerns and is our first direct measurement of privacy constructs. 

This is the first study we know of about framing effects on privacy judgments in a HRI. Just a few studies have been done in privacy-sensitive robotics; although two \citep{butler_privacy-utility_2015,hubers_video_2015} focus explicitly on remotely-operated robots, none focuses on the effects of framing or on the way the robot's actions appear from a third-person perspective instead of from its video feed. Past studies of framing in human-robot interactions focus on, e.g., user perceptions of the robot's social role and degree of anthropomorphism, but not privacy, and we are unaware of any framing studies of telepresence robots.

\section{Related Work}
\label{sec:related-work}
\subsection{Privacy}
Privacy is important in all human cultures \citep{altman_privacy_1977}, although different cultures have different norms for privacy and different mechanisms for enforcing those norms. Prominent theories that describe privacy include Altman's \citep{altman_environment_1975} and Nissenbaum's \citep{nissenbaum2004privacy}. We use a privacy taxonomy compiled from the literature \citep{rueben_taxonomy_2016} to divide ``privacy'' into component ideas: (1) Informational privacy, over personal information, includes (a) Invasion, (b) Collection, (c) Processing, and (d) Dissemination; (2) Physical privacy, over personal space or territory, includes (a) Personal Space, (b) Territoriality, and (c) Modesty; (3) Psychological privacy, over thoughts and values, includes (a) Interrogation and (b) Psychological Distance; (4) Social privacy, over interactions with others and influence from them, includes (a) Association, (b) Crowding/Isolation, (c) Surveillance, (d) Solitude, (e) Intimacy, (f) Anonymity, and (g) Reserve. We used this taxonomy to define the extents of the idea of privacy so we can work towards covering it all in our research program (i.e., towards \emph{content validity}). The small amount of work so far on ``privacy-sensitive robotics'' includes \citet{hubers_video_2015,butler_privacy-utility_2015}, and \citet{rueben_interfaces_2016}.

\subsection{Privacy Concerns about Telepresence Systems}

Robot-mediated communication has become possible for doctors, workers, bosses, and visitors to older adults \citep{markoff_boss_2010}. These telepresence robots create interactions that differ both from face-to-face interactions \citep{boyle_privacy_2009} and from purely virtual systems like avatar-based telepresence \citep[e.g.,][]{bente_social_2004,bente_effects_2007,lee_now_2011}. Even video media spaces change the privacy situation because the remote operator's actions are seen outside his/her context and dissociated from his/her identity \citep{boyle_privacy_2009}. Giving the remote user a physical (robot) body raises additional normative questions, like whether it is acceptable to rest one's feet on the robot's base \citep{lee_now_2011}. Several studies on telepresence robots for older adults have identified privacy concerns \citep{beer_mobile_2011} and behavior changes due to feeling watched by the robot \citep{caine_effect_2012}. The paradigm shift expected from the advent of telepresence robots may even prompt changes to U.S. privacy law \citep{kaminski_robots_2015}. 
%
This appears the be the first study that focuses on both privacy concerns and telepresence robots (in Section~\ref{sec:study-4}). 

%
%

\subsection{Animations for Studying HRI}
Several studies have compared human-robot interactions over video to live ones. \citet{woods_methodological_2006} have shown a strong agreement in general between being approached by a real robot and by a robot in a video. The same authors also cite findings by developmental psychologists, however, that ``while babies happily interact with their mothers via live video, they get highly distressed when watching pre-recorded or replayed videos of their mothers (as it lacks the contingency between mother's and baby's behaviour)''  \citep{woods_comparing_2006}. We were careful not to use our videos to study scenarios that normally require interaction. 

\citet{powers_comparing_2007} compared interactions with an animated computer agent against interactions with a real robot over a video feed. They found that engagement was higher and positive personality traits were more strongly associated with the real robot, but also that people remembered what the agent said better than what the robot said. Also, \citet{mcdonnell_evaluating_2008} compared emotive actions performed by a human actor to the same actions mapped onto animated bodies. They found that the perception of emotions in the actions was mostly the same across body conditions. 

Our methodology resembles the one used by \citet{takayama_expressing_2011}, which prototypes robot behaviors as animations and shows the videos to a large sample of people online. We use Amazon Mechanical Turk (MTurk) to recruit our subjects---consult \citet{mason_conducting_2012} for some studies that compare MTurk users to laboratory subjects. Performing a study with animated robot behaviors allowed us to provide consistent experiences for each of the study participants, to test a variety of task domains, and to engage a geographically diverse set of study participants. In terms of design research, using animations allowed us to test the behaviors we would like to build before locking in the design a robot would need to physically perform the behaviors. 

\subsection{Framing}
A frame is a ``structure of expectation'' \citep{tannen_whats_1993} within which actions and words will be interpreted differently. Framing language is metacommunicative; it tells one the frame in which to interpret subsequent communications \citep{bateson_theory_1955}. \citet{bateson_theory_1955} gives the example of monkeys engaged in a playful fight; the monkeys know that the bite that would normally be aggressive is fun in this context. Here the frame is ``play'', and ``this is `play''' would be framing language. \citet{tannen_whats_1993} reviews the idea of a ``frame" as well as the related terms ``schema'' and ``script'' across disciplines.  

\citet{groom_responses_2011} cite some studies of existing expectations about robots, but observe that ``researchers have largely ignored studying framing as an independent variable.'' Their study manipulates the role of a robot, an instance of framing, in a search and rescue context. \citet{howley_effects_2014} also manipulate the robot's social role, whereas \citet{fischer_initiating_2014} have a robot issue a greeting to frame the interaction as social. \citet{paepcke_judging_2010} manipulate user expectations about the robot's capabilities, a framing that is very relevant to the concerns of \citet{richards_how_2012}. \citet{darling_whos_2015} uses a narrative about the robot to manipulate how much users anthropomorphize it. We consider all these  studies to use framing. The studies presented in this paper use framing to manipulate the familiarity of the robot operator to the participant via short operator biographies in (uniquely) a telepresence scenario.

\section{Approach}
\label{sec:approach}
\begin{figure}
\centering
\includegraphics[width=0.85\columnwidth]{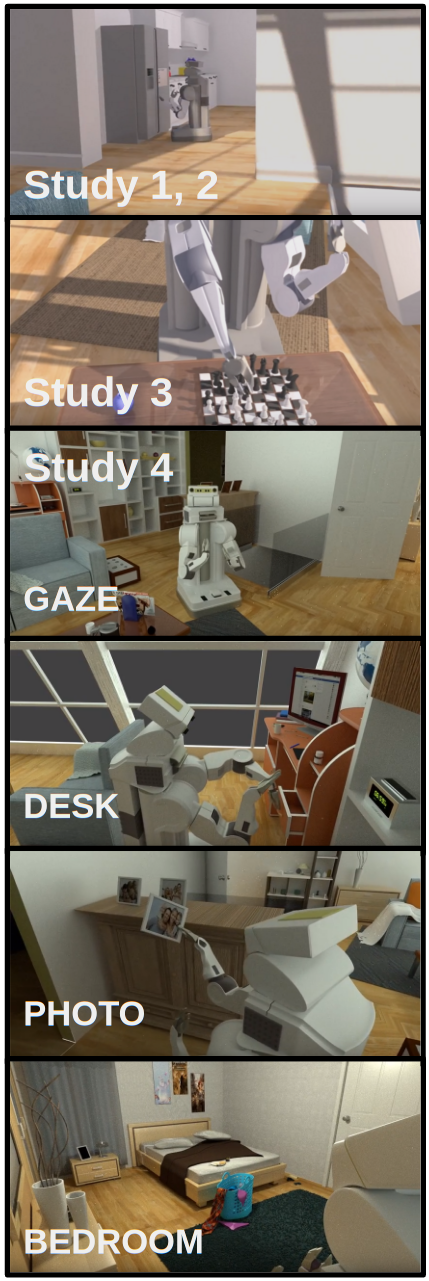}
\caption{Representative thumbnails from the animated videos used for the studies presented in this paper. See Sections~\ref{sec:study-1}--\ref{sec:study-4} for video descriptions.}
\label{fig:thumbnails}
\end{figure}

We use the same five-part methodology for each of the four studies reported here. 

(1) We \emph{frame} our participants by having them read a paragraph of text. This text describes the occasion of the interaction and introduces the robot operator, whose identity we manipulate in our experiments. Our goal was to provide enough context for the video to make sense but leaving the robot's actions ambiguous enough to require interpretation. 

(2) Next, we present our main \emph{stimulus}, which is an animated video of a PR2 robot performing some actions inside a home (see Figure~\ref{fig:thumbnails}). We chose the PR2 because it has a mobile base and two arms for performing human-like manipulation tasks as well as a head and obvious eyes so users can follow its gaze, but lacks a screen for showing imagery of the remote operator---something that would introduce many variables we don't want to deal with yet. The animated videos were made in Blender. We chose to use animation so we could specify the layout and appearance of the home, and also so that the robot could perform natural-looking actions that are not only beyond the state-of-the-art for autonomy but also difficult to do using teleoperation. 


(3) \emph{Responses} are gathered using a questionnaire. We expect subjects to interpret the videos from within the contextual frame we've given them and respond accordingly. 

(4) We \emph{interpret} what is being measured by the survey responses using principal component analysis (PCA). 

(5) We \emph{iterate} on this process by using the results from one study to re-design our survey for the next study. Successive studies can act as (at least partial) replications that contribute to a meta-analysis. Iterating is made easy by the response speed on Amazon Mechanical Turk; our studies typically complete within a few hours of being launched. Also, all the elements of our studies are easy to change quickly: a paragraph for framing, an animated video for the main stimulus, and a questionnaire for the response.

\section{Study 1: Opening the Fridge}
\label{sec:study-1}

Our first study tested whether the framing effect is measurable, and if so how large it is (RQ1-A). In all four studies our two frames manipulate the subject's familiarity with the robot operator within the hypothetical scenario.

\subsection{Methods}

\emph{Frame.} This study used a between-groups design with 2 framing conditions. The manipulated variable was the familiarity of the respondent's relationship with the robot operator in the hypothetical scenario. The operator was ``your sister'' in one framing and ``a home appraiser'' whom ``you have spoken with{\dots}once over the phone, but have never met{\dots}before in person'' in the other. In both cases we used the name Lisa. A general motivation was provided for each character so the scenario was not confusing: the sister is seeing your remodeled kitchen for the first time, whereas the home appraiser is checking that it's up to code.  

\emph{Main Stimulus.} The video (16s long) shows an animated PR2 robot in the kitchen. The robot opens the refrigerator, looks inside, and closes it again. The camera perspective is from the next room over, the living room.

\emph{Survey Items.} We created 7 items about trust, comfort, and acceptability to measure respondents' concerns about each scenario. They all used a 7-pt Likert-type response format. Two open-ended questions checked for attention and understanding of the two scenarios. No demographic information was collected in the survey.

We recruited 64 people total---31 saw the ``sister'' condition and 33 the ``stranger'' (i.e., home appraiser) condition. Subjects in each of the 4 studies were paid approximately \$10 per hour (\$1.50 per HIT for this study based on a predicted duration of about 9 minutes) and we always ensured that nobody took the same survey twice.

\subsection{Results}
Data reduction via principal component analysis (PCA) yielded 2 dependent variables: 6 items were combined whereas 1 remained separate, namely, ``I think it's important for me to be at home while this is happening.'' The 6-item composite we named Comfort (\(\alpha\) = .95); the other item we will call ShouldBeHome for short. Correlation between these two dependent variables was \textit{r}(62) = -.37.  

Respondents were more comfortable and felt less need to stay home in the ``sister'' condition (RQ1-A). Two-sample t-tests showed these framing effects to be statistically significant with medium-high effect sizes (on Comfort: \textit{t}(62) = 3.39, \textit{d} = 0.847, \textit{p} = 0.00123; on ShouldBeHome: \textit{t}(62) = -3.16, \textit{d} = -0.790, \textit{p} = 0.00245). 

Many people described the home appraiser's behavior as ``nosy'' in the open-ended responses. We decided to use this word in future versions of the survey because it fits the sorts of concerns we tried to evoke in our scenario; apparently many respondents also found it apt. 

%
%

\subsection{Discussion}
These initial results show that the framing effect is present, sizable, and able to be detected via an online survey with text prompts and an animated video. The next 2 studies measure additional variables while replicating and generalizing the framing effect.

\section{Study 2: Fridge (W/in Subjects)}
\label{sec:study-2}
The second study aimed to test what happens when a person switches between interpretive frames. We did this by re-framing our subjects in the middle of the experiment, resulting in a repeated measures design. First, we sought to replicate the framing effect we detected in Study 1 (RQ2-A). Second, we wondered whether showing both frames in series would reveal any carry-over effects, such as a halo effect that would make judgments more similar or a contrast effect that would make them more different (RQ2-B). 

\subsection{Methods}
In this study, each subject saw both framing conditions. The framings and video were the same as for Study 1 except that the home appraiser is now a home ``inspector'' named ``Alice'' (your sister is still named ``Lisa''). The order of conditions was counterbalanced. In between the two conditions we showed a text note that told respondents that an entirely new scenario will now be presented, but with the same video.  

\emph{Survey Items.} We used the same questions as in Study 1 with some minor wording changes as well as the addition of two new questions about whether Lisa was being ``nosy'' or ``rude,'' inspired by some open-ended responses in Study 1. We also moved the open-ended items from the end of the questionnaire to its beginning to encourage respondents to think about what they saw before responding to the other items. We added a more explicit manipulation check after both conditions---``In one sentence, describe the difference between the two scenarios''---to make sure respondents read and understood the framings.

We analyzed responses from 41 people after omitting 5 for failing the manipulation checks---of the 41, 22 saw ``sister'' first and 19 saw ``stranger'' first. Eight had also participated in Study 1.

\subsection{Results}
Unlike for Study 1, data reduction here yielded a single dependent variable (\(\alpha\) = .95). We named this single, 9-item dependent variable Comfort, but it consisted of different items than in Study 1. This data reduction remained stable even when responses were split by condition or order except for one interesting exception: ShouldBeHome was more strongly correlated with BeingRude and BeingNosy in the ``stranger'' condition than in the ``sister'' condition. 

Respondents were more comfortable with the ``sister'' condition (RQ2-A). According to an ANOVA, this within-subjects framing effect on Comfort was statistically significant with a large effect size (\textit{F}(1,40) = 18.4, \textit{\(\eta_p^2\)} = .315, \textit{p} < .001). We also checked for an order effect on Comfort, but it was not statistically significant and was much smaller in magnitude (\textit{F}(1,40) = 1.34, \textit{\(\eta_p^2\)} = .033, \textit{n.s.}). 

Regarding the carry-over effect, subjects' Comfort ratings were more sensitive to framing when the ``sister'' condition was viewed first (RQ2-B). A two-sample t-test on the differences in means between the two framing conditions, however, revealed that this effect was not statistically significant (mean difference in Comfort for ``sister'' first: 1.73, for ``stranger'' first: 0.98, \textit{t}(39) = 1.16, \textit{p} = .253). 

\subsection{Discussion}
This study successfully replicated the large framing effect from Study 1. The effect size may be different depending on which condition comes first, but more evidence is needed to confirm this. We continue counterbalancing the order of conditions in our subsequent studies to prevent order from confounding our framing manipulation, as well as to continue measuring any difference in effect size based on which frame is presented first.
%
%

\section{Study 3: Playing Chess}
\label{sec:study-3}
We next test whether the framing effect generalizes to a different video and scenario description (RQ3-A). We also conduct a second test of whether the framing effect is moderated by which frame comes first (RQ3-B) and a first test of whether the name of the operator matters (RQ3-C). Finally, we report on the range of subjects' sensitivity to the change in frames (RQ3-D) and test whether some basic demographics are correlated with our dependent variables (RQ3-E), 

\subsection{Methods}
\emph{Main Stimulus.} The video we used for this study shows a new scenario. The PR2 is now with the observer in the living room; it drives around a chess board to face the observer, looks down at a chess board, and makes a move. 

\emph{Frame.} The framing paragraphs that precede the video were rewritten to fit the new scenario, but still manipulated subjects' hypothetical familiarity with the operator: the operator is either a friend you play chess with frequently or a stranger you've only met on a chess website. Also, we counterbalanced the assignment of the operator's name (Lisa or Alice) to each framing condition (sister and stranger). 

\emph{Survey Items.} After each showing of the video, 17 items with Likert-type response formats were presented (paraphrased in Figure~\ref{fig:study-3-factors}). Of these, 7 were adapted from the 9 items in Study 2; 10 were new, many of which were tailored to this scenario (e.g., ``I think using a robot like this is a good idea for improving the online chess experience.''). Here we began using a 9-pt response format instead of a 7-pt one to combat floor and ceiling effects. 

Next we asked 20 demographics questions. These included general information such as age, sex, and education level as well as items about experience with robots and living situation (e.g., ``I am used to sharing my living space with other people such as friends, family, roommates, and guests.''). 

The 14 NARS items by \citet{nomura_experimental_2006} were presented next, but they were adapted to be about telepresence robots (e.g., ``If robots had emotions, I would be able to make friends with them'' became ``If robots could express emotions for someone far away, I feel I could become friends with that person'') and we clarified some wordings to address the comments by \citet{syrdal_negative_2009}. We will refer to these items with ``NATS'', which stands for ``Negative Attitudes about Telepresence (robots) Scale''.  

We analyzed 61 out of the 66 responses after excluding 5 people for failing our manipulation checks. 

\subsection{Results}
This sample was comprised of 38 men, 22 women, and 1 person who left that question blank. Most respondents (72\%) were white. Mean age was 32.5 years (SD: 9.2 years). Everyone had completed high school and only 4 were unemployed or unable to work. 89\% had at least seen a video of a robot before, and 43\% had driven a remote controlled vehicle. 75\% reported that they use social media daily and 85\% that they use a smartphone. Only 36\% live alone and only 16\% live with children. 

\begin{figure}
\centering
\includegraphics[width=0.70\columnwidth]{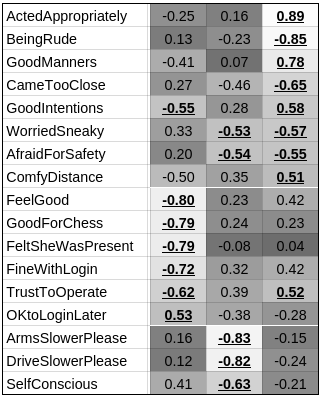}
\caption{Table of factor loadings for data reduction of 17-item video response survey in Study 3. From left to right, we named these PositivePresence, WasNotTooFast, and PilotEtiquette. Items placed in each composite are \underline{underlined}.}
\label{fig:study-3-factors}
\end{figure}

We chose to reduce our 17-item video response survey into 3 dependent variables named PositivePresence (7 items, \(\alpha\) = .92), WasNotTooFast (5 items, \(\alpha\) = .86), and PilotEtiquette (9 items, \(\alpha\) = .94). Figure~\ref{fig:study-3-factors} shows the items included in each variable and their factor loadings. Correlations between these variables range from \textit{r}(120) = +.69 to +.84.

All three variables were higher in the ``friend'' condition than in the ``stranger'' condition (RQ3-A). An ANOVA revealed that these framing effects were large and statistically significant (PositivePresence: \textit{F}(1,57) = 88.2, \textit{\(\eta_p^2\)} = .607, \textit{p} < .001; WasNotTooFast: \textit{F}(1,57) = 79.9, \textit{\(\eta_p^2\)} = .584, \textit{p} < .001; PilotEtiquette: \textit{F}(1,57) = 86.8, \textit{\(\eta_p^2\)} = .604, \textit{p} < .001). 

As in Study 2, the framing effect was larger when the ``friend'' (previously ``sister'') framing came before the ``stranger'' framing (RQ3-B). Yet again, however, this effect was not statistically significant (it was largest for WasNotTooFast: difference in means for ``friend'' first = 1.58, for ``stranger'' first = 1.07, \textit{t}(59) = 1.47, \textit{p} = .146).

Statistically significant name effects were found for PilotEtiquette (\textit{F}(1,57) = 5.82, \textit{\(\eta_p^2\)} = .093, \textit{p} = .019) and almost for WasNotTooFast (\textit{F}(1,57) = 3.54, \textit{\(\eta_p^2\)} = .058, \textit{p} = .065), but these turned out to be due to a breakdown of random assignment\footnote{Subjects assigned to the group with Lisa as the friend responded more affirmatively to the item, ``I really dislike it when people enter my bedroom without my permission.'' This imbalance mattered because subjects who responded more affirmatively to that item were also more sensitive to the framing manipulation. Thus, part of the framing effect appeared to be a naming effect. Adding that demographic variable to the ANOVA caused all naming effects to lose statistical significance.} (RQ3-C). None of the three order effects was statistically significant (all \textit{p}s > .1) and observed effect sizes were relatively small (all \textit{\(\eta_p^2\)}s < .04). 

There were notable individual differences in sensitivity to the framing manipulation (RQ3-D). For example, PositivePresence ratings changed by a mean of 1.53 points between framing conditions, but a few respondents changed not at all or in the opposite direction. Upon inspection, it looks like many of these people did not demonstrate their understanding of the conditions very well in the open-ended questions. Manipulation checks like these are crucial for gauging experimental validity in online surveys. Most people were affected by the framing in the expected way, however; the 95\% confidence interval for the mean sensitivity of PositivePresence only spans from 1.20 to 1.86 points. 

We kept 13 of the 14 items in the NATS and call it the NATS-13 (\(\alpha\) = .91). The last item in the scale, which reads, ``I feel that in the future society will be dominated by robots'', only correlated at \textit{r}(59) = .19 with the rest of the scale. Since it seems to measure a belief that could be relevant to judgments about our scenario, however, we chose to keep it as a single-item variable called SocietyWillBeDominated.

The NATS-13 was especially strongly correlated with our three dependent variables (PositivePresence \textit{r}(59) = -.71, \textit{\(r^2\)} = .50; WasNotTooFast \textit{r}(59) = -.73, \textit{\(r^2\)} = .53; PilotEtiquette \textit{r}(59) = -.71, \textit{\(r^2\)} = .50), accounting for half of their variance (RQ3-E). Actually, this was a stronger effect than the framing manipulation itself; moving up one point on the NATS-13 was about equivalent to switching the frame from ``friend'' to ``stranger''. These correlations were statistically significant even after a conservative Bonferroni correction to account for the heightened risk of Type I error from checking the correlations between 17 of our demographic variables and each of the 3 dependent variables.  
 
Many of our other demographic variables besides the NATS-13 were also correlated with ratings of PositivePresence, WasNotTooFast, and PilotEtiquette (e.g., ``In general, I enjoy having guests over to where I live") (RQ3-E). A few were correlated with sensitivity to the framing manipulation (e.g., ``I really dislike it when people enter my bedroom without my permission''). It is interesting that these two things don't always occur together (e.g., the ``enjoy having guests'' example above was not strongly correlated with sensitivity to the framing manipulation). 

\subsection{Discussion}
Our results support the conclusion that the framing effect generalizes to the chess scenario. This study has also yielded much more information than the first two did, including demographic profiles of participants, correlations between demographics and our dependent variables, and first looks at sensitivity and effects of operator name. We now turn to our constructs of interest in the realm of privacy.

\section{Study 4: Tour before a Party}
\label{sec:study-4}
This study uses four videos made to evoke different privacy concerns on our taxonomy \citep{rueben_taxonomy_2016} as well as survey items to measure these privacy concerns. Our main research question is whether the framing effect replicates in this domain (RQ4-A). We are also interested in several types of order effects. First, do respondents get used to seeing a robot poking around their house after watching the four videos, causing privacy concerns to be lower in the next condition (RQ4-B)? Second, is the framing effect moderated by which frame comes first (RQ4-C)? Third, does a frame wear off as respondents watch the videos (RQ4-D)? After all, the videos do not show the operator's face or any other explicit cues that he is telepresent. 

We will also take another look at whether the demographics we have targeted are correlated with any of our privacy variables (RQ4-E), as well as at how sensitive these new variables are to the framing manipulation (RQ4-F). Also, we want to know how much of the framing effect is attributable to changes in trust of the operator (RQ4-G). Finally, we look at which of the four videos is most (or least) concerning with respect to privacy (RQ4-H).

\subsection{Methods}
\emph{Frame.} The scenario is that you have invited some people to a party in your home, but one person can't attend except by logging into the robot. That person is either a close friend whom you see weekly or a stranger whom you met for the first time today. We used male names this time: Will and Chris. 

\emph{Main Stimulus.} The main stimulus for this study was divided into four new animated videos with survey items presented after each video. This way, participants could answer questions directly after watching a certain part of the scenario. The videos were more photorealistic than those used in the previous three studies (see Figure~\ref{fig:thumbnails}). They range from 21--34s long. Each is designed to evoke certain privacy constructs in order to cover as much of the taxonomy \citep{rueben_taxonomy_2016} as possible: e.g., Surveillance and Psychological Distance in the ``gaze'' video (the robot makes eye contact with you), Invasion and Territoriality in the ``desk'' video (looks in your desk drawer), Anonymity in the ``photo'' video (picks up a family photo), and Modesty in the ``bedroom'' video (sees your messy bedroom, including some women's underwear).  

\emph{Survey Items.} We used all new items for the video response survey. These were designed to target the privacy constructs evoked by the videos, as well as ask some more general questions. All 22 items are paraphrased in Figure~\ref{fig:study-4-factors}. They all use a 9-pt Likert-type response format. Note that these items, as in the other studies, are context-sensitive; they might tap different constructs if we used them in a different scenario.

Immediately after reading each framing paragraph, participants took a 7-item trust questionnaire. Six items were adapted from the Specific Interpersonal Trust Scales (SITS-M and SITS-F) \citep{johnson-george_measurement_1982} and one item, shown last, we created: ``I trust [Will/Chris]''. 

The demographics questions were the same as in Study 3 except that the NATS was modified slightly and reduced to 8 items for brevity.

At the very end of the survey we asked our respondents to rank our four videos by how concerning they were with respect to privacy.

This study was counterbalanced for the order of the two framings, for the order of the four videos according to a Latin Square design, and for which name was assigned to which framing as in Study 3. We analyzed 65 responses after discarding one that failed the manipulation check. 

\subsection{Results}
Of the 65 responses we analyzed, 40 were males and 25 were females. 62\% marked ``white'' as their ethnicity. The mean age was 32.0 years (SD: 8.8 years). All our respondents indicated that they had graduated high school, and all but 18\% are (self-)employed. 49\% had at least driven a remote-controlled vehicle before. 63\% use social media every day, and all but 1 owned a smartphone. Only 29\% live alone; another 29\% live with children at least sometimes. 

\begin{figure}
\centering
\includegraphics[width=0.95\columnwidth]{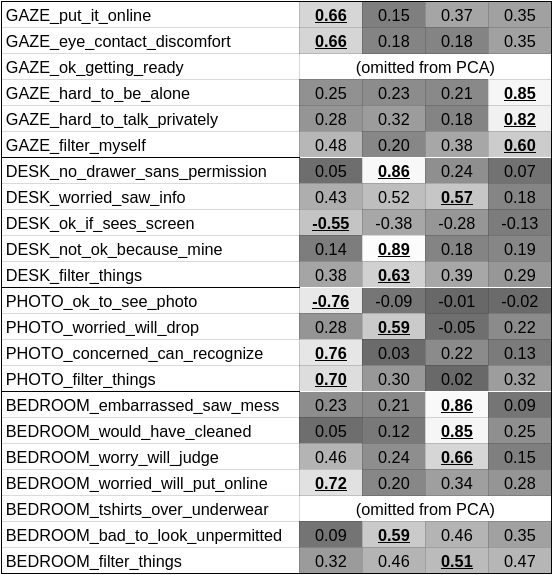}
\caption{Table of factor loadings for data reduction of 22-item video response survey in Study 4. From left to right, we named these WorriedAboutLikenesses, DontMessWithMyStuff, EmbarrassedByMess, and HardToBeAlone. Items placed in each composite are \underline{underlined}.}
\label{fig:study-4-factors}
\end{figure}

Data reduction was performed on all 22 video response questions at once to look at how the different privacy constructs might relate. Using a PCA we chose 4 composite variables for reporting (Figure~\ref{fig:study-4-factors}). Each composite variable combines items that were meant to tap different privacy constructs, indicating either that those constructs overlapped in our respondents' minds or that our items failed to discriminate between them. WorriedAboutLikenesses (7 items; \(\alpha\) = .89) includes items about dissemination of personal information, anonymity, and psychological distance; DontMessWithMyStuff (5 items; \(\alpha\) = .87) includes items about invasion of personal information and territory; EmbarrassedByMess (5 items; \(\alpha\) = .90) includes items about collection and processing of information, namely a messy room that could cause someone to judge you; HardToBeAlone (3 items; \(\alpha\) = .91) includes items about solitude, intimacy, and surveillance. Intercorrelations between composites ranged between \textit{r}(128) = +.60 and +.73. 

Two items, one designed to tap surveillance and one for modesty, didn't correlate much with any of the others (all \(|\textit{r}(128)|\) < .22). After analysis, we believe that they were not good measures of any relevant constructs to this study, so we do not report on them further here\footnote{Neither item had any statistically significant framing effects, which were very large for the rest of our dependent variables, or any statistically significant correlations with our demographic variables. The item texts were ``I would feel comfortable doing some small tasks to prepare for the party knowing that Will could drive by and look at me like this'' (``gaze'' video) and ``I wouldn't be as upset if the clothes basket contained t-shirts instead of underwear'' (``bedroom'' video).}.

The framing manipulation effect was always in the predicted direction for our four privacy variables: higher privacy concerns for a stranger than for a friend (RQ4-A). An ANOVA revealed that these effects were all large and statistically significant (WorriedAboutLikenesses: \textit{F}(1,49) = 91.8, \({\eta}_p^2\) = .652, \textit{p} < .001; DontMessWithMyStuff: \textit{F}(1,49) = 45.5, \({\eta}_p^2\) = .481, \textit{p} < .001; EmbarrassedByMess: \textit{F}(1,49) = 79.4, \({\eta}_p^2\) = .618, \textit{p} < .001; HardToBeAlone: \textit{F}(1,49) = 38.9, \({\eta}_p^2\) = .443, \textit{p} < .001).  

The effect of acclimatization (i.e., first condition vs. second condition) was only statistically significant for DontMessWithMyStuff (\textit{F}(1,49) = 4.71, \({\eta}_p^2\) = .088, \textit{p} = .035) and almost for HardToBeAlone (\textit{F}(1,49) = 3.04, \({\eta}_p^2\) = .058, \textit{p} = .088). These were both in the expected direction as well: concerns about the operator messing with your stuff or making it hard for you to be alone were lower in the second condition, suggesting that respondents became complacent as they got used to the scenarios (RQ4-B). The name (Will or Chris) of the operator was also included in the model, but was not found to have any statistically significant effects.

Just like in Studies 2 and 3 we checked whether sensitivity to framing was moderated by which frame came first. There was a consistent effect: the relationship effect was larger on all 4 composites when the respondent saw the ``stranger'' frame first (RQ4-C). This is in the opposite direction, however, as it was in Studies 2 and 3. To test for statistical significance, we ran two-sample t-tests for each of the 4 composites between two groups: those who saw ``friend'' first (n = 31) and those who saw ``stranger'' first (n = 34). This difference was only statistically significant for DontMessWithMyStuff (mean sensitivity for ``friend'' first = -0.77, for ``stranger'' first = -1.56, \textit{t}(63) = 2.41, \textit{d} = 0.601, \textit{p} = .019). 

In this study we tested whether the effects of a frame wear off as the respondents watch the four videos. For each video (e.g., ``desk'') within each frame (e.g., ``friend'') we ran a MANOVA (8 total) that tested whether that video's position in the order of videos (e.g., 3rd of 4) predicts the responses to the items that go with that video. None of these 8 tests was statistically significant; it appears that our two frames remained active for the extent of each condition, and perhaps could have for much longer, or until a significant distraction occurred (RQ4-D). 

We chose 5 out of the 8 NATS items (the ``NATS-5''; \(\alpha\) = .81) for a measure of negative emotions about telepresence systems. The remaining three we kept as single-item variables. The first was ``If robots could express emotions for someone far away, I feel I could become friends with that person'' (correlations with the other 3 variables were all \(|\textit{r}(63)| < .43\)). The other two were written to figure out why the item that said ``I feel that in the future society will be dominated by robots'' didn't fit into the rest of the scale in Study 3; one reads, ``I feel that in the future robots will be everywhere in our society'', and the other reads, ``I feel that in the future society will be controlled by robots.'' The ``society will be controlled'' item appears to be the misfit: it was not correlated with the NATS-5 (\textit{r}(63) = -.04). The ``robots will be everywhere'' item was negatively correlated (\textit{r}(63) = -.46) with the NATS-5, so it appears that discomfort with telepresence was linked in our sample with a belief that robots will not become ubiquitous anytime soon.

Some of our demographic variables were correlated with our privacy DVs (RQ4-E). The largest correlation was \textit{r}(128) = +.40 between a composite of CleanUpBeforeGuests and CloseDoorsBeforeGuests (\(\alpha\) = .49) and EmbarrassedByMess. Also, the NATS-5 was correlated with WorriedAboutLikenesses at \textit{r}(128) = +.36 and also the other 3 main composites, but those correlations---and, in fact, all other IV-DV correlations---were not statistically significant when we protect for Type I error from checking all possible correlations. Note that the NATS items account for much less variance here than in Study 3, wherein the 13-item NATS composite accounted for half of the variance of the dependent variables. 

We also looked at how sensitive our respondents' privacy concerns were to the relationship manipulation (RQ4-F). In this study, the variable with the most individual differences in sensitivity was HardToBeAlone---its 95\% confidence interval spanned from a sensitivity of -1.99 to -1.02. None of the confidence intervals for our 4 composite privacy variables crossed zero. There were also some sizable correlations between demographic variables and these sensitivities, especially with the living situation and NATS variables, but none reached statistical significance after protecting for Type I error. 

According to a PCA, the 7-item scale about trusting the operator measured a single dimension, which we called ``Trust'' (\(\alpha\) = .95). It is interesting to note that although the SITS \citep{johnson-george_measurement_1982} from which we drew all but 1 of the items claims to measure multiple dimensions of trust, with slightly different dimensions between men and women, our items only appear to have measured one construct for both sexes in this study. We propose that this is because the respondents do not know much about the robot operator in either framing condition, so their judgments about him are not as complex as they might be for, e.g., a close friend in real life.

We can validate our composite measure of Trust by testing whether it is manipulated by the difference in frames (``friend'' vs. ``stranger'') and not by the difference in names (``Will'' vs. ``Chris''). Matched-pair t-tests supported the validity of our Trust measure (relationship effect: \textit{t}(64) = 13.2, \textit{d} = 3.31, \textit{p} < .0001; name effect: \textit{t}(64) = 0.349, \textit{d} = 0.871, \textit{p} = 0.729). 

Higher Trust ratings do appear linked to decreased privacy concerns (RQ4-G). Correlations are statistically significant with WorriedAboutLikenesses (\textit{r}(128) = -.50) and HardToBeAlone (\textit{r}(128) = -.35). So Trust is a significant mediator of the effect of hypothetical familiarity on privacy concerns, but there are probably other significant ones to be discovered: at most Trust accounts for 25\% of the variance (WorriedAboutLikenesses) or at least a mere 7.5\% (DontMessWithMyStuff and EmbarrassedByMess). 

Respondents ranked the ``desk'' video as most concerning (votes: most 45---12---6---2 least), followed by ``bedroom'' (most 14---43---6---2 least), then ``photo'' (most 4---4---33---24 least), then ``gaze'' (most 2---6---20---37 least) (RQ4-H). It makes sense with our gut intuition because ``desk'' and ``bedroom'' are more invasive, we think, than mere ``gaze'' or touching a ``photo''.

\section{Discussion}
\label{sec:discussion}
\subsection{Implications for Design}
All four studies show a large effect of framing the relationship of the local user with the robot operator. This suggests that robot designers should think about how frames like these could be encouraged via robot appearance, what the robot says, and how the robot is advertised. Designers should also consider how different contexts and cultural factors, like the popular media and even other people, could impose unwanted frames over a user's interaction with a robot.

We are beginning to understand some details of how framing works. We see, starting in Study 2, that people can be prompted to switch between interpretive frames relatively easily. Some frames may ``stick'' better than others, though, even changing the way framing information is interpreted. Study 4 suggests that certain privacy concerns decrease as time passes or as people experience different frames. We believe people simply get used to the concerning behaviors and lower their guards, although an alternative explanation is that experimental realism begins to decrease with the second framing. An understanding of how privacy concerns about robots change with long-term usage is crucial; if concerns wear off or change in type after a few hours or days then robots will need to transition smoothly between these two phases.  


\vfill\eject

\subsection{On Methodology}
It is surprising that the framing effect was large even with \emph{no} signs of the operator's presence beyond the framing text. It was even present (and large) between subjects, when there were no hints about a contrast between the two frames. We would hypothesize that adding signs of the operator's presence would prolong or even increase framing effects as long as they agree with the frame. On the other hand, we are interested in what happens when conflicting information about the frame is presented to an observer, e.g., if the observer is told the robot is being used to inspect a leaky pipe but instead starts picking up your personal items.

We chose a methodology that uses animated video stimuli, quantitative data, and quick, online recruitment. We believe the animated videos are useful for studying experiences that are difficult to produce in the laboratory, which includes anything from natural robot gestures to scenarios in outer space! Also, shorter studies can operate like pilot studies when aspects of the methodology are not well-established, helping you fix problems before wasting months on a large, one-off experiment. Similarly, taking quantitative data yields early effect size estimates so we can choose proper sample sizes. 

Some qualitative methods can also be short and easy to iterate on, such as focus groups. Showing our framings and videos to a focus group would be another way to explore which variables are important. Although we would not get effect size estimates or the type of insight into relationships between variables offered by PCA, a focus group conversation could explore a much broader selection of concerns and use nuanced follow-up questions to greatly increase our confidence that participants' responses mean what we think they mean. 

\subsection{Future Work}
\label{sec:future-work}
Future work should concentrate on which frames have the largest effect sizes on privacy concerns. We have only looked at the difference between a familiar person and a stranger operating the robot in these first 4 studies. Here are some other aspects of the frame one could manipulate: the level of control the operator has over the robot's action (full teleoperation vs. supervision); how invasive the operator's actions are expected to be based on his/her role (police officer vs. tourist); morphology of the robot; robot nonverbal behavior (e.g., smoothness of movements, posture); or the operator's social presence (via a face on a screen or operator name tag). Knowing which aspects of the context most influence user perception of privacy risks \citep[as in][]{hancock_meta-analysis_2011} will help researchers and robot manufacturers know what to focus on. 

The framing effect itself should also be studied. How is the frame encoded and then used to interpret subsequent information? How much longer does it last beyond the four videos and 22 survey items in Study 4? What happens, exactly, when people switch framings? Also, studying the individual differences (e.g., personality traits) that moderate framing effects would help identify groups that are especially sensitive.

We are motivated by privacy. Understanding framing will help us avoid privacy violations. More research could help us discover how privacy judgments are different in HRI; we want to identify what type(s) of privacy exactly will be problematic and which people to target with our solutions.

\renewcommand*{\bibfont}{\raggedright}


\end{document}